\documentclass[conference, a4paper]{IEEEtran}
\IEEEoverridecommandlockouts
\usepackage{svg}
\usepackage{multirow}
\usepackage{amsmath,amssymb,amsfonts,mathtools}
\usepackage{placeins}
\usepackage{algorithm,algpseudocode}
\usepackage{orcidlink} 
\usepackage[capitalise, noabbrev]{cleveref}
\usepackage[nolist]{acronym}
\usepackage{balance}

\usetikzlibrary{arrows.meta, positioning, calc}

\usepackage{siunitx}
\usepackage{pifont}
\usepackage{subcaption}
\usepackage{graphicx}
\algrenewcommand\algorithmiccomment[1]{\hfill /*#1*/}

\usepackage{pgfplots}
\pgfplotsset{compat=1.18}
\usepackage{filecontents}

\usepackage{balance}
\usepackage{mleftright}

\def\BibTeX{{\rm B\kern-.05em{\sc i\kern-.025em b}\kern-.08em
    T\kern-.1667em\lower.7ex\hbox{E}\kern-.125emX}}

\bstctlcite{IEEEexample:BSTcontrol}

\makeatletter
\newcounter{problem}
\newcounter{tmp}

\makeatother

\usepackage{listings}
\usepackage{color} 
\definecolor{myblue}{HTML}{00549f}
\definecolor{myred}{HTML}{a11035}
\definecolor{mypurple}{HTML}{7a6fac}
\definecolor{myorange}{HTML}{f6a800}
\definecolor{mygreen}{HTML}{57ab27}
\definecolor{myviolett}{HTML}{612158}

\usepackage{enumitem}

\usepackage{stfloats}

\definecolor{deepblue}{rgb}{0,0,0.5}
\definecolor{deepred}{rgb}{0.6,0,0}
\definecolor{deepgreen}{rgb}{0,0.5,0}
\definecolor{backcolour}{rgb}{0.95,0.95,0.92}

\crefname{lstlisting}{Listing}{Listings}
\Crefname{lstlisting}{Listing}{Listings}

\usepackage[all]{background}
\usepackage{stackengine}
\setstackEOL{\\}
\setstackgap{L}{\normalbaselineskip}
\SetBgContents{\color{gray}{\tiny \Longstack{
    PREPRINT - Accepted for publication at the Design, Automation \& Test in Europe Conference \& Exhibition (DATE), April 20-22, 2026, in Verona, Italy. \\
    \textbf{© 2026 IEEE:} Personal use of this material is permitted. Permission from IEEE must be obtained for all other uses, \\ including reprinting/republishing this material for advertising or promotional purposes, collecting new collected works for resale \\ or redistribution to servers or lists, or reuse of any copyrighted component of this work in other works.
}}}
\SetBgPosition{4.5cm,1cm}
\SetBgOpacity{1.0}
\SetBgAngle{0}
\SetBgScale{1.8}

\begin{document}
\bstctlcite{IEEEexample:BSTcontrol}

\begin{acronym}
    \acro{rram}[RRAM]{Resistive Random Access Memory}
    \acro{cim}[CIM]{Computing-in-Memory}
    \acro{mvm}[MVM]{Matrix-Vector Multiplication}
    \acro{cnn}[CNN]{Convolutional Neural Network}
    \acro{nn}[NN]{Neural Network}
    \acro{ofm}[OFM]{Output Feature Map}
    \acro{ifm}[IFM]{Input Feature Map}
    \acro{gpu}[GPU]{Graphics Processing Unit}
    \acro{cpu}[CPU]{Central Processing Unit}
    \acro{tpu}[TPU]{Tensor Processing Unit}
    \acro{ml}[ML]{Machine Learning}
    \acro{gpeu}[GPEU]{General Purpose Execution Unit}
    \acro{pe}[PE]{Processing Element}
    \acro{dac}[DAC]{Digital-to-Analog Converter}
    \acro{qat}[QAT]{Quantization-Aware Training}
    \acro{wbs}[WBS]{Weight Bit Slicing}
    \acro{ibs}[IBS]{Input Bit Slicing}
    \acro{mpq}[MPQ]{Mixed-Precision Quantization}
    \acro{onnx}[ONNX]{Open Neural Network Exchange}
    \acro{haq}[HAQ]{Hardware-aware Automated Quantization}
    \acro{cim-aq}[CIM-AQ]{CIM-aware Automated Quantization}
    \acro{gemm}[GEMM]{General Matrix Multiplication}
    \acro{1t1r}[1T1R]{1 Transistor 1 Resistor}
    \acro{ddpg}[DDPG]{Deep Deterministic Policy Gradient}
    \acro{api}[API]{Application Programming Interface}
    \acro{qdq}[QDQ]{Quantize and DeQuantize}
    \acro{tvm}[TVM]{Tensor Virtual Machine}
    \acro{adc}[ADC]{Analog-to-Digital Converter}
    \acro{mlp}[MLP]{Multi-Layer Perceptron}
    \acro{mha}[MHA]{Multi-Head Attention}
\end{acronym}

\title{
Mixed-Precision Training and Compilation for RRAM-based Computing-in-Memory Accelerators
}

\def\finalpaper{1}

\if\finalpaper1
{
    \author{
    \IEEEauthorblockN{Rebecca Pelke$^1$\IEEEauthorrefmark{1},
        Joel Klein$^1$\IEEEauthorrefmark{1},\\
        José Cubero-Cascante\IEEEauthorrefmark{1}, Nils Bosbach\IEEEauthorrefmark{1},
        Jan Moritz Joseph\IEEEauthorrefmark{1}\IEEEauthorrefmark{2}, Rainer Leupers\IEEEauthorrefmark{1}}
    \IEEEauthorblockA{
        \IEEEauthorrefmark{1}RWTH Aachen University, Germany\\
        \IEEEauthorrefmark{2}RooflineAI GmbH, Germany\\
        \IEEEauthorrefmark{1}\{pelke, klein, cubero, bosbach, joseph, leupers\}@ice.rwth-aachen.de\hspace{0.3cm}
        \IEEEauthorrefmark{2}joseph@roofline.ai}
    \thanks{\vspace{-0.2cm}$^1$Both authors contributed equally to this work.}
    \vspace{-0.5cm}
    }
}
\else
    \author{
      \IEEEauthorblockN{Authors are removed for submission}
      \\
      \\
      \IEEEauthorblockA{Affiliations are removed for submission}
      \thanks{Authors are removed for submission}
      \\
    }
\fi

\maketitle

\DeclareSIUnit{\bits}{bits}

\begin{abstract}
Computing-in-Memory (CIM) accelerators are a promising solution for accelerating \ac{ml} workloads,
as they perform \acp{mvm} on crossbar arrays directly in memory.
Although the bit widths of the crossbar inputs and cells are very limited,
most CIM compilers do not support quantization below \SI{8}{\bit}.
As a result, a single MVM requires many compute cycles,
and weights cannot be efficiently stored in a single crossbar cell.

To address this problem, we propose a mixed-precision training and compilation framework for \acs{cim} architectures.
The biggest challenge is the massive search space, that makes it difficult to find good quantization parameters.
This is why we introduce a reinforcement learning-based strategy to find suitable quantization configurations that balance latency and accuracy.
%
In the best case, our approach achieves up to a $\mathbf{2.48\times}$ speedup over existing state-of-the-art solutions,
with an accuracy loss of only \SI{0.086}{\percent}.
\end{abstract}

\begin{IEEEkeywords}
    RRAM, CIM, Compiler, ML, MPQ
\end{IEEEkeywords}
\acresetall 

\vspace{-0.2cm}
\section{Introduction}

To perform \ac{ml} workloads efficiently, new hardware architectures are required.
Promising candidates are \ac{cim} accelerators, which execute \acp{mvm} directly in memory.
This addresses the von Neumann bottleneck by reducing data movements between memory and processing units.~\cite{amrouch2021towards}

\ac{rram} is a well-known technology for \ac{cim} crossbars because it offers high device density,
low power usage, fast switching speed,
and compatibility with standard CMOS processes~\cite{zahoor2020resistive,wu2022a}.

A limitation of \ac{rram} crossbars is the restricted bit precision of both the input activations and the crossbar cells.
The cell resolution is limited by device variability, nonlinearity, and drift~\cite{grossi2015impact,ding2023bnn,pelke2025show}.
This makes it difficult to reliably distinguish between many resistance states.
Typical bit widths for \ac{rram} cells are between \SI{1}{\bit} and \SI{4}{\bits}~\cite{xue201924,wan2022compute,read2025neurosim,pelke2024lats}.
Input resolution is limited by the area, power consumption, and speed of the \ac{dac} used to convert the digital inputs to analog voltages~\cite{wei2022trends,ni2016energy}.
Higher-resolution \acp{dac} demand higher-resolution \acp{adc}, which increases power consumption significantly~\cite{wan2022compute}.
As a result, most crossbars use \SI{1}{\bit} \acp{dac} to reduce
circuit complexity and improve robustness and efficiency~\cite{chou2019cascade,bengel2023bit}.

Existing compilers for \ac{cim} only support fixed bit width quantization with a minimum of \SI{8}{\bit} for activations and weights~\cite{drebes2020tc,vadivel2020tdo,siemieniuk2021occ,Jin.2023,khan2024cinm,qu2024cim}.
To map such \SI{8}{\bit} models onto low-resolution crossbars, two techniques are used:
\textit{weight bit slicing} and \textit{input bit slicing}~\cite{xiao2021accuracy}.
Weight bit slicing splits high-bit weights across multiple \ac{rram} cells.
Input bit slicing splits high-bit inputs across multiple computing cycles.
This increases the inference latency drastically~\cite{xiao2021accuracy}.

Mixed-precision \ac{qat} with low bit widths is a hardware-friendly alternative to fixed-precision quantization~\cite{huang2021mixed}.
It combines \ac{qat}~\cite{jacob2018quantization},
which takes simulation effects into account during training,
and \ac{mpq}, where different layers use different bit widths for weights and inputs.
To make use of this quantization scheme on \ac{cim} targets,
we present our framework shown in \Cref{fig:overview}.
It has two main contributions:

\begin{figure}[!tp]
  \centering
  \includegraphics[width=\linewidth, keepaspectratio]{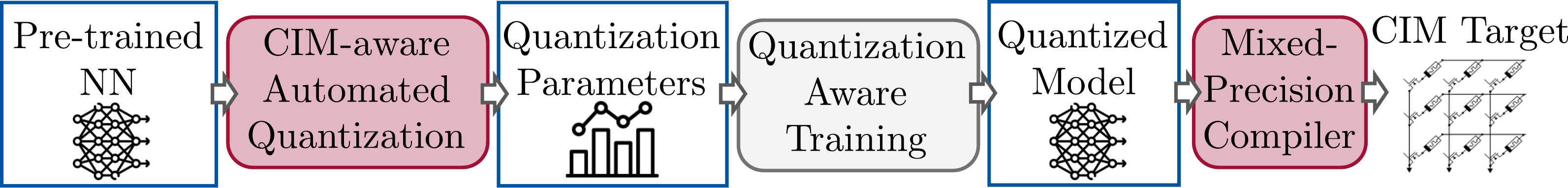}
  \caption{Overview of the proposed framework.
  The main contributions are highlighted in red.}
  \label{fig:overview}
\end{figure}

\textbf{1. A MPQ-aware compiler}
that compiles mixed-precision models from the \ac{qat} framework Brevitas~\cite{alessandro_pappalardo_2025_15375017} for \ac{cim} architectures.
The supported architectures are described in \Cref{sec:related_work,sec:mvm}.
It can handle crossbars of any size and applies \ac{cim}-specific optimizations.
The compiler is described in \Cref{sec:cim_compiler}.
Results show speedups ranging from $2.20\times$ to $2.48\times$
over \SI{8}{\bit} quantization compilers for ResNet-18, ViT-B/32, and VGG-16 on the ImageNet dataset.

\textbf{2. A reinforcement learning-guided MPQ optimizer}
for \ac{cim} targets.
The exponential search space for \ac{mpq} makes manual tuning impractical.
Existing automated approaches~\cite{wang2019haq,dong2019hawq} target only conventional hardware,
while open-source frameworks for \ac{cim} are missing.
We close this gap with \ac{cim-aq}, our \ac{cim}-aware reinforcement learning-based optimizer.
Therefore, we extend the \ac{haq} framework~\cite{wang2019haq} to enable mixed-precision optimization for \ac{cim}.
By integrating Brevitas~\cite{alessandro_pappalardo_2025_15375017} as a new \ac{qat} backend,
\ac{cim-aq} does not only support a broader range of \acp{nn}, including transformers,
but is also approximately \SI{8}{\percent} faster.
\ac{cim-aq} can be found on GitHub\footnote[2]{\vspace{-0.2cm}GitHub link: \url{https://github.com/jmkle/cim-aq}}.


\section{Related Work}
\label{sec:related_work}

\begin{table}[!tbp]
    \centering
    \footnotesize
    \caption{\label{tab:relatedwork}Comparison with existing \ac{cim} compilers.}
    \begin{tabular}{c|cccc}
    Compiler &
    \begin{tabular}[c]{@{}c@{}}Cell\\resolution\end{tabular} &
    \begin{tabular}[c]{@{}c@{}}Crossbar\\size\end{tabular} &
    \begin{tabular}[c]{@{}c@{}}Data\\types\end{tabular} &
    \begin{tabular}[c]{@{}c@{}}MPQ\\support\end{tabular}
    \\ \hline
    TC-CIM~\cite{drebes2020tc} & \SI{4}{\bit} & $256\times 256$ & \SI{8}{\bit} & \ding{55} \\
    TDO-CIM~\cite{vadivel2020tdo} & \SI{4}{\bit} & $256\times 256$ & \SI{8}{\bit} & \ding{55} \\
    OCC~\cite{siemieniuk2021occ} & \SI{4}{\bit} & variable & \SI{8}{\bit} &\ding{55} \\
    Jin et al.~\cite{Jin.2023} & \SI{2}{\bit} & variable & \SI{16}{\bit} & \ding{55} \\
    CINM~\cite{khan2024cinm} & \SI{4}{\bit} & variable & \SI{8}{\bit} &\ding{55} \\
    CIM-MLC~\cite{qu2024cim} & any & variable & \SI{8}{\bit}& \ding{55} \\
    \hline
    Ours & any & variable & \SI{1}{\bit}-\SI{8}{\bit} & \ding{51}
    \end{tabular}
    \vspace{-0.4cm}
\end{table}

\begin{figure}[!tp]
  \centering
  \includegraphics[width=\linewidth, keepaspectratio]{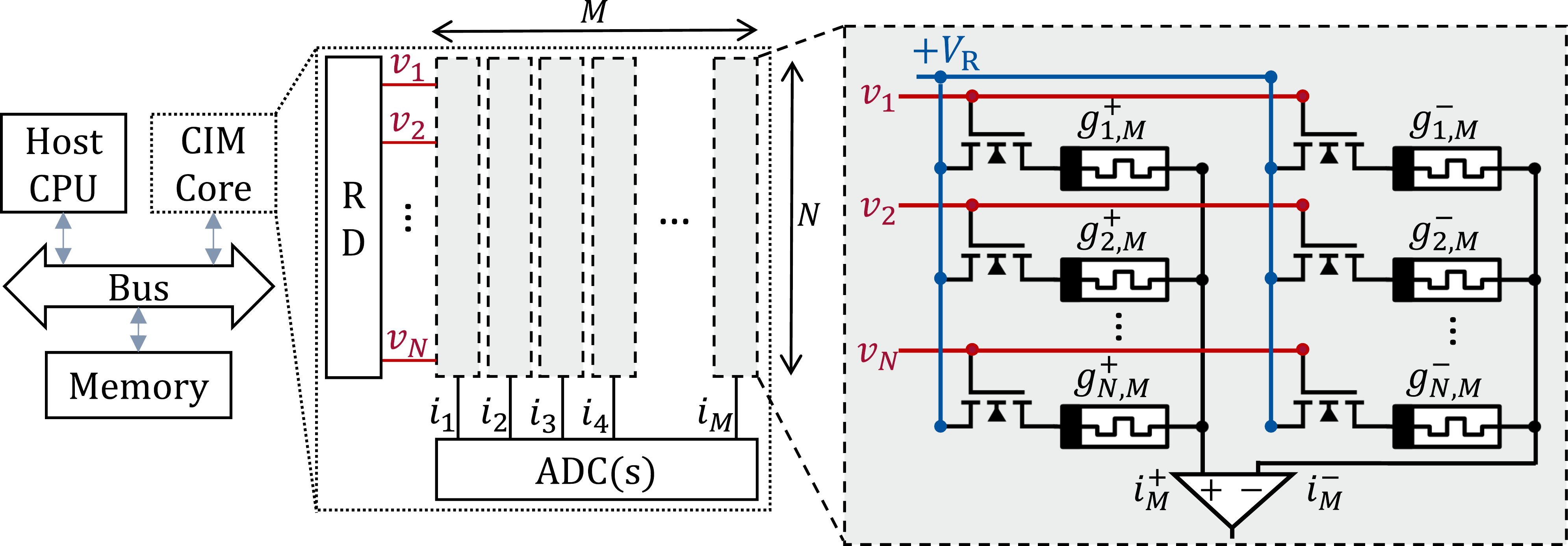}
  \caption{The \ac{cim} architecture used in this work.}
  \label{fig:architecture}
  \vspace{-0.4cm}
\end{figure}

Many \ac{cim} compilers have been proposed in the past~\cite{drebes2020tc,vadivel2020tdo,siemieniuk2021occ,Jin.2023,khan2024cinm,qu2024cim,Ambrosi.2018,Han.2022}.
They differ in the targeted \ac{cim} architecture, crossbar design, compiler framework, implemented optimizations, and overall flexibility.
\Cref{tab:relatedwork} shows compilers that assume a \ac{cim} architecture similar to ours.
\Cref{fig:architecture} illustrates this architecture, consisting of a host CPU and a single memory-mapped \ac{cim} accelerator.
Besides the crossbar, the \ac{cim} core also includes registers and control logic, which are omitted here for clarity.

TC-CIM~\cite{drebes2020tc} uses Tensor Comprehensions~\cite{vasilache2019next} and Loop Tactics~\cite{chelini2019declarative}
to detect and offload suitable tensor operations like \acp{mvm} and \acp{gemm} to a \ac{cim} accelerator.
TDO-CIM~\cite{vadivel2020tdo} builds on TC-CIM by detecting patterns at the LLVM-IR level
for broader language support and uses Polly~\cite{grosser2012polly} and Loop Tactics to offload individual layers to a \ac{cim} accelerator.
OCC~\cite{siemieniuk2021occ} uses MLIR~\cite{lattner2021mlir} to offload \acp{gemm} and convolutions to a \ac{cim} accelerator.
They improve endurance through reduced writes and better weight reuse.
Jin et al.~\cite{Jin.2023} developed a general-purpose LLVM-based compiler that identifies \acp{mvm}, \acp{gemm},
and logic operations, with a runtime application that decides between CPU and CIM execution.
CINM~\cite{khan2024cinm} offers an end-to-end compiler for heterogeneous CIM systems,
using hierarchical MLIR abstractions for progressive lowering and optimizations.
CIM-MLC~\cite{qu2024cim} proposes a multi-level compilation stack with progressive
lowering and scheduling optimizations tailored to different CIM architecture levels.

As summarized in \Cref{tab:relatedwork},
current compilers are restricted to a fixed bit width of \SI{8}{\bit} or \SI{16}{\bit}.
Since most crossbars have only \SI{2}{\bit} to \SI{4}{\bit} resolution and \SI{1}{\bit} \acp{dac},
the overhead of each \ac{mvm} becomes large because multiple cycles per \ac{mvm} and many cells per weight are required.
As a result, latency increases and efficiency drops.
%
To address these issues, our compiler uses the best bit width for each layer.
This reduces cycles and write operations and thereby improves latency.

\section{Background}
This section provides background on \ac{rram}-based \ac{cim} architectures, \ac{qat},
and the used \ac{haq} framework.

\subsection{Analog MVMs on RRAM Crossbars}
\label{sec:mvm}
\ac{rram} crossbars are used to perform \acp{mvm} directly in memory.
\Cref{fig:architecture} illustrates the basic architecture of the \ac{cim} target
used in this work.
The topology of the $M\times N$ \mbox{\ac{1t1r}} crossbar
has a high resilience against wire parasitics~\cite{xiao2021analysis,cubero2025evaluating}.
Each column consists of $2N$ cells.
Each pair of cells represents a single weight; $g_{j,M}^+$
is the positive and $g_{j,M}^-$ the negative part of the weight.
This is called \textit{differential mapping}~\cite{xiao2021accuracy,pelke2025optimizingbinaryternaryneural}.
The dot product result of a crossbar column $i_{k}$, with $k \in \mleft[1,M\mright]$, can be written as:
\vspace{-0.2cm}
\begin{align}
i_{k} = i_k^+ - i_k^- = \sum_{j=1}^{N} v_j \cdot \mleft(g_{j,k}^+ - g_{j,k}^-\mright)
\end{align}
\vspace{-0.2cm}

The input $v_j$ is used to activate or deactivate row $j$.
The input is binary and requires one cycle per bit.
Each pair of cells can store a multi-bit weight.
Weight bit slicing is used if the weight has more bits than the cell can store.

\subsection{Quantization-Aware Training}
Quantization in \ac{ml} usually means mapping floating point ranges to integer values,
e.g., \texttt{fp32} ranges to \texttt{int8} values.
In \textit{range-based linear quantization}, this mapping is described by
a \textit{scaling factor} $s$ and an offset $z$ called \textit{zero-point}.

In our setup, weights use \textit{symmetric} signed quantization,
and activations use symmetric signed or unsigned quantization.
In symmetric quantization, the zero-point is set to zero.
The symmetric quantization of a floating point value $x_\mathrm{f}$ to an integer $x_\mathrm{q} \in [-\mleft(2^{B-1}-1 \mright),2^{B-1}-1]$ is defined as:
\vspace{-0.2cm}
\begin{align}
x_\mathrm{q} = \mleft\lfloor x_\mathrm{f} \cdot \frac{2^{B-1} - 1}{\max \mleft(|x_\mathrm{f}|\mright)} \mright\rceil
\end{align}
\vspace{-0.2cm}

For small bit widths $B$, the quantization error can be significant.
To improve the accuracy, \ac{qat}~\cite{jacob2018quantization} is commonly used.
An important concept in \ac{qat} is \textit{fake quantization}.
Fake quantization applies rounding and clipping to weights and activations in the forward pass
but lets gradients pass through unchanged, so the \ac{nn} learns the quantization effects.

A well-known framework for \ac{qat} is Brevitas~\cite{alessandro_pappalardo_2025_15375017}.
Brevitas is based on PyTorch and also supports \ac{mpq}.

\subsection{Reinforcement Learning}
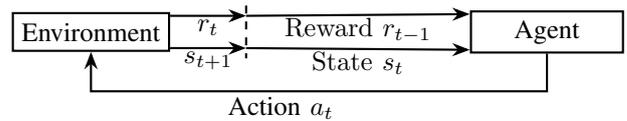
\begin{figure}[!bp]
  \vspace{-0.5cm}
  \centering
    \begin{tikzpicture}[
    box/.style={rectangle, draw, minimum width=2cm, minimum height=0.5cm, align=center},
    every node/.style={font=\normalsize},
    ->, >=Stealth, thick
    ]
    \node[box] (env) at (0,0) {Environment};
    \node[box] (agent) at (6,0) {Agent};
    \node[coordinate] (top) at ($(env.east) + (1cm,0.35cm)$) {};
    \node[coordinate] (bottom) at ($(env.east) + (1cm,-0.35cm)$) {};
    \draw[-, dashed] (top) -- (bottom);
    \draw[->] ($(env.south east)+(0,0.05cm)$) -- ++(1,0) node[below, xshift=-0.5cm, yshift=0.08cm] {$s_{t+1}$};
    \draw[->] ($(env.north east)+(0,-0.05cm)$) -- ++(1,0) node[below, xshift=-0.5cm, yshift=0.08cm] {$r_{t}$};
    \draw[->] ($(env.south east)+(1,0.05cm)$) -- node[below, yshift=0.09cm] {$\mathrm{State} \ s_{t}$} ($(agent.south west)+(0,0.05cm)$);
    \draw[->] ($(env.north east)+(1,-0.05cm)$) -- node[below, yshift=0.09cm] {$\mathrm{Reward} \ r_{t-1}$} ($(agent.north west)+(0,-0.05cm)$);
    \draw[->] (agent.south) -- ++(0,-0.5) -- ++(-6,0) -- node[below, xshift=2.5cm, yshift=-0.2cm] {Action $a_t$} (env.south);
    \end{tikzpicture}
  \vspace{-0.25cm}
  \caption{Agent-environment interaction.}
  \label{fig:ReinforcementLearning}
  \vspace{-0.2cm}
\end{figure}

Reinforcement learning~\cite{kaelbling1996reinforcement} is a concept where an \textit{agent} learns by interacting with an \textit{environment} over time
through \textit{actions} and \textit{rewards}.
\Cref{fig:ReinforcementLearning} illustrates the agent-environment interaction.
%
At each time step $t$, the agent chooses an action $a_t$ according
to a \textit{policy}~$\pi$, with $a_t = \pi (s_t)$ for deterministic policies~\cite{silver2014deterministic}
and $\pi (a_t | s_t)$ for stochastic policies~\cite{chou2017improving}.
The environment responds with a new state $s_{t+1}$ and a reward $r_{t}$.

The goal in deterministic reinforcement learning is to find a policy~$\pi^*$ that maximizes the expected cumulative reward:
\vspace{-0.1cm}
\begin{align}
\pi^*(s) = \arg\max_{\pi} \ \mathbb{E} \mleft[ \sum_{k=0}^{\infty} \gamma^k \, r_{t+k} \,\middle|\, s_t = s \mright],
\end{align}
where $\gamma \in [0,1)$ is a discount factor that prioritizes rewards.

\subsection{\acl{haq} (HAQ)}
\label{sec:haq_framework}
\begin{figure}[!tp]
  \centering
  \includegraphics[width=\linewidth, keepaspectratio]{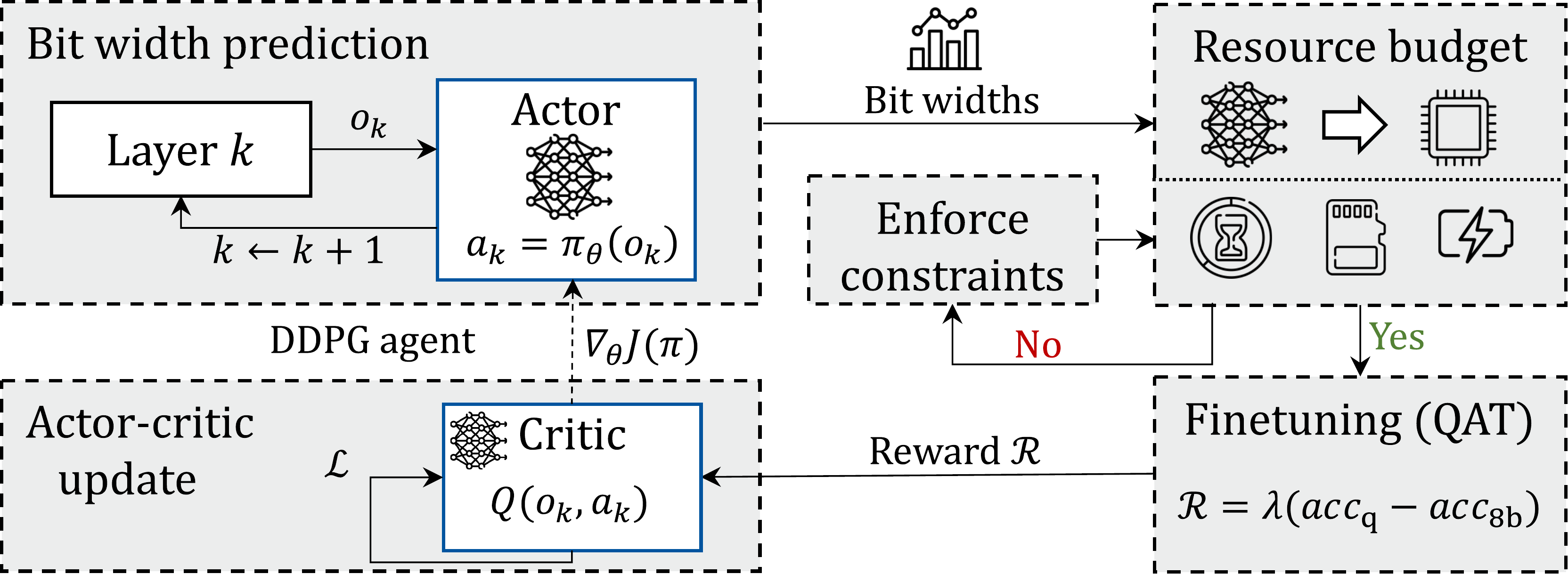}
  \caption{Learning flow in the \ac{haq} framework.}
  \label{fig:haq}
  \vspace{-0.4cm}
\end{figure}

\vspace{-0.1cm}
The HAQ framework uses a \ac{ddpg}~\cite{tan2021reinforcement} agent.
\ac{ddpg} is an \textit{off-policy} algorithm,
which means that the agent can learn from data generated by a different policy than the one it is currently optimizing.
It is based on the actor-critic architecture~\cite{grondman2012survey}, 
where the \textit{actor} learns the policy~$\pi$
and the \textit{critic} learns the action-value function~$Q(s,a)$.
In \ac{haq}, a state is denoted as $o_k$ and contains the layer information of layer $k$.
The policy and the action-value function are learned by \acp{nn}.
\Cref{fig:haq} breaks down the main steps of the learning flow of the \ac{haq} framework:
\begin{enumerate}
\item In each \textit{episode}, the actor determines the quantization parameters $a_k=\pi_{\theta} \mleft(o_k\mright)$ for each layer separately.
\item It is checked if the quantized \ac{nn} fits into the resource budget for latency, memory, and power consumption.
\item If the resource budget is exceeded, the bit widths are decreased sequentially to enforce the constraints.
\item The \ac{nn} is trained for only one epoch. The reward $\mathcal{R}$ is calculated based on the top-1 accuracy of this epoch.
\item All tuples $T_k=\mleft(o_k, a_k, \mathcal{R}, o_{k+1}\mright)$ are stored in the \textit{replay buffer}, which collects previous experiences.
\item The critic network is trained by minimizing a loss based on a variant of the Bellman equation~\cite{bellman1952theory}. Therefore, a batch of random samples from the input buffer is used. The actor is updated using gradients from the critic.
\end{enumerate}

After training, a full \ac{qat} must be performed using the learned parameters.
We will use Brevitas for this step, as its trained \ac{nn} can be directly imported into the compiler.

\section{Our Approach}
\vspace{-0.1cm}
\label{sec:implementation}
\Cref{sec:rl_adaptions} describes the \ac{cim-aq} framework,
which adapts the \ac{haq} framework to \ac{cim} targets
and accelerates it.
\Cref{sec:cim_compiler} introduces the \ac{mpq}-capable compiler.

\subsection{CIM-aware Automated Quantization (CIM-AQ)}
\label{sec:rl_adaptions}
In HAQ, the agent searches for the \ac{mpq} policy that maximizes accuracy while staying within hardware's resource budgets
(memory footprint, latency, power consumption).

\sisetup{parse-numbers = false}
In \ac{cim}, however, we need to solve the following problem:
Find \ac{mpq} parameters that \textbf{minimize latency} and
\textbf{preserve high accuracy} for a given \ac{cim} target.
This leads to the following adjustments of the \ac{haq} framework:

\subsubsection*{\textbf{Reward function}}
The target accuracy $acc_\mathrm{t}$ that should at least be achieved by the \ac{mpq} is defined as
\vspace{-0.1cm}
\begin{align}
    acc_\mathrm{t} = acc_\mathrm{8b} - acc_\mathrm{loss}
\end{align}
\vspace{-0.0cm}
while $acc_\mathrm{8b}$ is the accuracy of the reference \SI{8}{\bit} \ac{nn} and $acc_\mathrm{loss}$
is the maximum tolerated accuracy loss.
Unlike the fixed hardware‑budget constraints in \ac{haq},
our target accuracy $acc_\mathrm{t}$ cannot be enforced as a strict constraint,
since the post‑quantization accuracy, $acc_\mathrm{q}$, of the \ac{mpq}‑quantized \ac{nn} is unknown at the time of constraint evaluation
(see \Cref{fig:haq}).
Instead, we incorporate $acc_\mathrm{t}$ into the reward function to penalize any accuracy loss below the target,
thereby enabling a joint optimization of latency and accuracy.

Our reward function $\mathcal{R}$ is defined as:
\begin{align}
\mathcal{R} &= 
    \begin{cases}
    -\alpha \mleft( acc_\mathrm{t} - acc_\mathrm{q} \mright), & acc_\mathrm{q} < acc_\mathrm{t} \\
    \beta \mleft( \frac{T_{\mathrm{8b}}}{T_{\mathrm{q}}} - 1\mright) + \gamma \mleft( acc_\mathrm{q} - acc_\mathrm{t} \mright), & \mathrm{otherwise}
    \end{cases}
\label{eq:reward_function}
\end{align}

The inference latencies of the \SI{8}{\bit} and \ac{mpq} quantized \acp{nn} are denoted as $T_{\mathrm{8b}}$ and $T_{\mathrm{q}}$, respectively.
The hyperparameter $\alpha$ scales the penalty when accuracy falls below the target,
whereas $\beta$ and $\gamma$ scale the trade-off between speedup and accuracy.
They are set to $\alpha=10$, $\beta=100$, and $\gamma=0.1$.

\subsubsection*{\textbf{Latency cost modeling}}
\begin{table}[!tp]
    \centering
    \footnotesize
    \caption{\label{tab:cim_hw_params}Hardware parameters of the CIM target.}
    \vspace{-0.1cm}
    \begin{tabular}{c|c}
    Parameter &
    Description
    \\ \hline
    $M \times N$ & Crossbar dimension, as shown in \Cref{fig:architecture} \\
    $r_{\mathrm{cell}}$ & Precision of a single \ac{rram} cell (in bit) \\
    $r_{\mathrm{DAC}}$  & Precision of the \ac{dac} (in bit) \\       
    $t_{\mathrm{write}}$ & Programming time of the crossbar (in \si{\micro\second}) \\
    $t_{\mathrm{mvm}}$ & Execution time of an \ac{mvm} (in \si{\micro\second})
    \end{tabular}
    \vspace{-0.4cm}
\end{table}
The inference latency $T$ of the \ac{nn} is determined based on the hardware parameters listed in \Cref{tab:cim_hw_params}.
These parameters belong to the architecture discussed in \Cref{sec:mvm}
with a single \ac{cim} core.
The total latency $T$ can be calculated as follows:
\begin{align}
T = \sum_{\mathrm{layer}~l} r_{\mathrm{repeat},l} \mleft( N_{\mathrm{write},l} \cdot t_{\mathrm{write}} + N_{\mathrm{mvm},l} \cdot t_{\mathrm{mvm}} \mright)
\end{align}
%
%
The repeat factor $r_{\mathrm{repeat},l}$ denotes the repetition factor for operations executed multiple times,
such as attention head computations in \ac{mha} layers.

The number of write operations $N_{\mathrm{write},l}$ is defined as:
\vspace{-0.1cm}
\begin{align}
N_{\mathrm{write},l} =
\mleft\lceil \frac{2\cdot M_{l}}{M} \cdot \mleft\lceil \frac{w_{\mathrm{bit},l}}{r_{\mathrm{cell}}} \mright\rceil \mright\rceil
\cdot \mleft\lceil \frac{N_{l}}{N} \mright\rceil
\end{align}
%
Here, $w_{\mathrm{bit},l}$ denotes the weight bit-width in layer $l$.
$M_l$ and $N_l$ are the row and column dimensions of the layer's weight matrix.
Layers without a 2D weight matrix, e.g., Conv2D, must be transformed into a \ac{gemm} operation first.
This is usually done with the \textit{im2col} transformation~\cite{zhang2020efficient,pelke2023Mapping,pelke2024clsa}.
For instance, a Conv2D layer mapping $(C_{\mathrm{in}}, H_{\mathrm{in}}, W_{\mathrm{in}}) \rightarrow (C_{\mathrm{out}}, H_{\mathrm{out}}, W_{\mathrm{out}})$
with kernel size $(K_{\mathrm{H}}, K_{\mathrm{W}})$
can be expressed as a $(M_l, V_l, N_l)$-\ac{gemm} operation with dimensions:
\vspace{-0.0cm}
\begin{align}
    M_l = C_{\mathrm{out}}, \quad
    N_l = C_{\mathrm{in}}  K_{\mathrm{H}} K_{\mathrm{W}}, \quad
    V_l = H_{\mathrm{out}} W_{\mathrm{out}}
    \label{eq:gemm_dims}
\end{align}
%

The number of \ac{mvm} operations $N_{\mathrm{mvm},l}$ is defined as:
\vspace{-0.0cm}
\begin{align}
N_{\mathrm{mvm},l} = V_{l} \cdot
N_{\mathrm{write},l}\cdot \mleft\lceil \frac{a_{\mathrm{bit},l}}{r_{\mathrm{DAC}}} \mright\rceil
\end{align}
The activation bit width is $a_{\mathrm{bit},l}$.
\Cref{eq:gemm_dims} specifies $V_l$ for Conv2D layers.
For normal Dense layers, we can simply set $V_l=1$.
For Dense and \ac{gemm} layers inside \ac{mha} blocks,
each token needs to be processed independently, so we set $V_l$ to the sequence length $N$.
%
We also generate a lookup table for the cost model.
It stores the latency for each layer
and every combination of weight and activation bit widths
in a range from $b_{\mathrm{min}}$ to $b_{\mathrm{max}}$, the minimum and maximum bit widths.

\subsubsection*{\textbf{HAQ acceleration}}
In the \ac{haq} framework, one epoch of \ac{qat} is performed during every finetuning phase
(see \Cref{fig:haq}).
One problem is that \ac{haq} relies on a custom quantization implementation,
which only supports very few layers.
For example, typical layers of transformer architectures are not supported.
Another problem is that the time of the whole \ac{haq} framework is dominated by the \ac{qat} epoch.
To overcome these limitations, we integrate Brevitas as a backend for \ac{qat} in the \ac{haq} framework.
Brevitas not only supports a wide range of layers, but also speeds up the whole reinforcement learning process.

\subsection{CIM Compiler}
\label{sec:cim_compiler}
\begin{figure}[!tp]
  \centering
  \includegraphics[width=\linewidth, keepaspectratio]{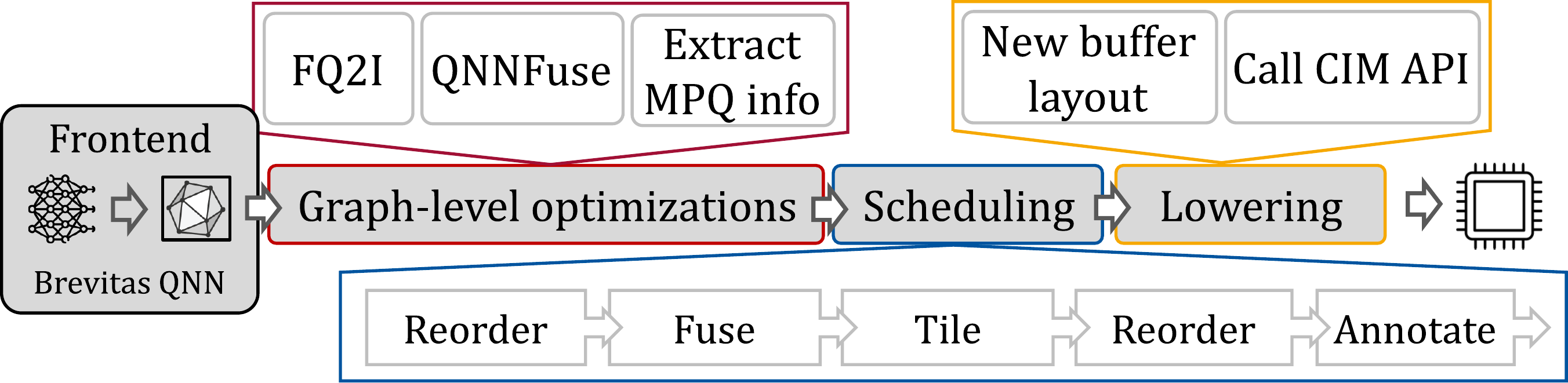}
  \caption{Overview of the compilation pipeline.}
  \label{fig:compiler}
  \vspace{-0.4cm}
\end{figure}

Our compiler maps \ac{mpq} models trained with Brevitas
to the \ac{cim} targets described in \Cref{sec:mvm}.
CIM-specific \ac{api} calls are inserted into the \ac{ml} model.
A runtime environment implements those calls to offload MVM executions to the CIM accelerator.
The compiler is built on the \ac{tvm}~\cite{chen2018tvm} framework
to reuse existing infrastructure.
~\Cref{fig:compiler} shows an overview of our compilation pipeline.
In the following, we will explain the main steps of the compilation pipeline.

\subsubsection*{\textbf{Frontend}}
After \ac{qat} with Brevitas, the \ac{mpq} model is converted to the \ac{onnx} format.
\ac{tvm} offers a built-in \ac{onnx} frontend.
The \ac{onnx} graph is then converted into an internal representation called \textit{Relay}.
Quantization is encoded in the \ac{qdq} style:
\texttt{QuantizeLinear} followed by \texttt{DeQuantizeLinear} nodes are inserted between the original operators.
These nodes allow switching between integer and floating-point domains.
At this point, operations like Conv2D are still in the floating-point format.

To enable \ac{mpq} in our compiler,
we added a \texttt{config\_update} attribute to Relay operators,
and implemented a Relay transformation pass that detects quantized layers,
infers activation and weight bit-widths,
and determines the bit-splitting scheme for the crossbar.
The resulting configuration is stored as JSON in the operator attributes and later used during lowering to generate accelerator function calls.

\subsubsection*{\textbf{Graph-level optimizations}}
\begin{figure}[!tp]
  \centering
  \includegraphics[width=\linewidth, keepaspectratio]{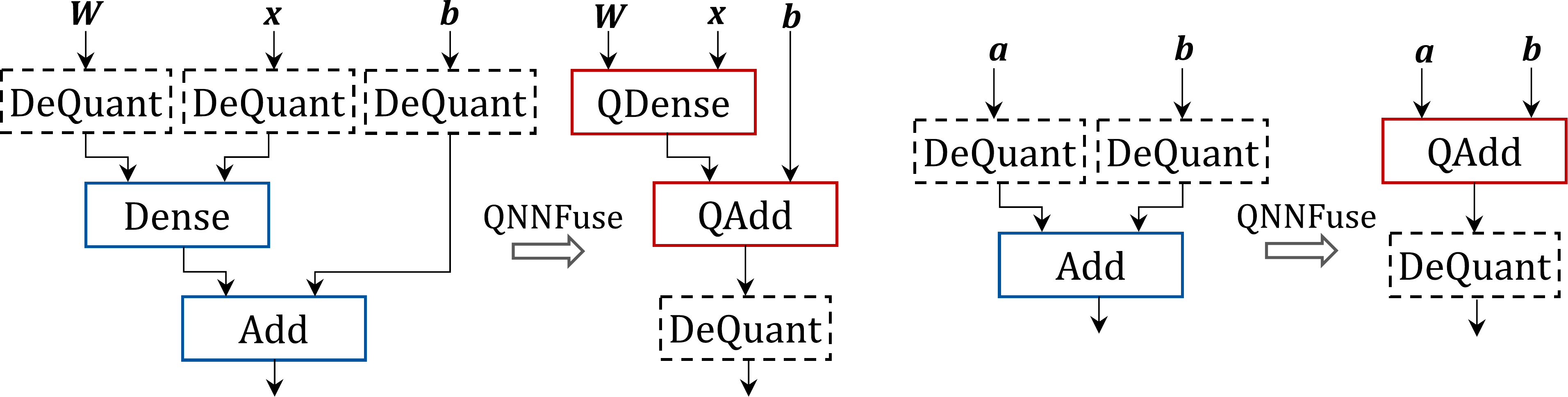}
  \caption{Optimizations of the QNNFuse pass.}
  \label{fig:qnnfuse}
  \vspace{-0.3cm}
\end{figure}

We apply a series of high-level transformations to the Relay graph.
For execution on the \ac{cim} target,
all \ac{qdq} patterns must be converted to integer-only computations.
TVM's \textit{FQ2I} pass replaces each \ac{qdq} pair with a quantized operator
and inserts \texttt{ReQuantize} nodes to align scales and zero points.
However, this pass can leave the final layer or the bias addition partly in floating-point format.
To ensure that the last block is also quantized, we implement a custom \textit{QNNFuse} pass.
It merges the leaf operations to corresponding quantized operations.
\Cref{fig:qnnfuse} illustrates two examples of graph transformations performed by this pass.

Finally, we partition the optimized graph,
assigning each operation either to the CPU or to the \ac{cim} accelerator.
CPU-only parts are lowered through the standard \ac{tvm} pipeline,
whereas \ac{cim} operations follow the custom scheduling and lowering flow shown in \Cref{fig:compiler}.

\subsubsection*{\textbf{Scheduling}}
Scheduling in TVM controls how an operation's loop nests are transformed to match a given hardware target.
Primitives such as \texttt{split}, \texttt{reorder}, \texttt{tile}, and \texttt{fuse}
(see \Cref{fig:compiler}) help to modify loop nests to improve parallelism and data locality.
%
Since write operations to the crossbar are more costly than \acp{mvm},
we adopt a weight-stationary dataflow that keeps weights in the crossbar as long as possible.
To target \ac{cim}, loops are \textit{annotated} with a predefined set of axis labels
(e.g., ``outer \ac{mvm} row'').
The idea is to only label the relevant loops in the schedule,
so our subsequent lowering passes pick up those labels to insert the proper CIM API calls fully automatically.
With this approach, supporting new layers only needs 5-10 lines of code.

\subsubsection*{\textbf{Lowering}}
We provide two custom lowering passes.
First, we insert staging buffers of size $M \times N$, $1\times N$,
and $1\times M$ to hold the weights, inputs, and outputs, respectively.
This aligns the memory layout with the required format of the CIM API.
Then, we inject API calls that replace the isolated computation in the inner loop nest.
Pointers to the buffers are passed as arguments.
\vspace{-0.3cm}


\section{Results}
In this section, we evaluate the performance of our framework.
In summary, the workflow consists of three steps:
First, reinforcement learning is employed to determine suitable quantization parameters.
Second, our compiler maps the optimized models to the CIM target
introduced in~\Cref{sec:mvm}.
Finally, the compiled model is executed on an open-source crossbar simulator~\cite{pelke2025optimizingbinaryternaryneural}.
The inference latency is determined by using crossbar parameters from~\cite{wan2022compute}:
The crossbar size is $256\times 256$ with differential cell mapping and $r_{\mathrm{DAC}}=1$.
The latencies are $t_{\mathrm{write}}=56\,\si{\micro\second}$ and $t_{\mathrm{mvm}}=1.4\,\si{\micro\second}$ (see \Cref{tab:cim_hw_params}).
%
These latencies also include the \ac{dac} and \ac{adc} conversions.
We optimize three representative \acp{nn} trained on ImageNet~\cite{deng2009imagenet}: ResNet-18~\cite{he2016resnet}, VGG-16~\cite{simonyan2015vgg}, and the vision transformer ViT-B/32~\cite{dosovitskiy2021vit}.
\Cref{tab:benchmarks} lists the number of layers that are mapped to the crossbar for each benchmark.

\begin{table}[!b]
    \vspace{-0.3cm}
    \centering
    \footnotesize
    \caption{\label{tab:benchmarks}An overview of the benchmarks' layers.}
    \begin{tabular}{c|ccc}
    Benchmark &
    Conv2D &
    Dense &
    MatMul
    \\ \hline
    ResNet-18 & 20 & 1 & - \\
    VGG-16 & 13 & 3 & - \\
    ViT-B/32 & 1 & 49 & 24
    \end{tabular}
    \vspace{-0.4cm}
\end{table}

\subsection{CIM-AQ - Performance Improvements}
To improve the \ac{haq} framework, we replaced the original quantization backend with Brevitas.
As mentioned in \Cref{sec:implementation}, this not only enables the use of a wider range of \ac{nn} architectures like transformers,
but also provides modest runtime improvements.
For example, on our NVIDIA L40S GPU (CUDA 12.8, 48 GB memory),
we observed a speedup of $\SI{8.4}{\percent}$ when running reinforcement learning for VGG-16 compared to the original HAQ implementation.

\begin{figure*}[!t]
    \centering
    \begin{tikzpicture}
    \begin{axis}[
        ybar,
        bar width=4pt,
        width=\textwidth,
        height=3cm,
        ylabel={\shortstack{Weight\\bit width}},
        xlabel={},
        symbolic x coords={Conv1,Conv2,Conv3,Conv4,Conv5,Conv6,Conv7,Conv8,Conv9,Conv10,Conv11,Conv12,Conv13,Conv14,Conv15,Conv16,Conv17,Conv18,Conv19,Conv20,Dense},
        xtick=\empty,
        x tick label style={rotate=45, anchor=east},
        nodes near coords,
        every node near coord/.append style={font=\small},
        legend style={
            at={(0.5,1.5)},
            anchor=north,
            legend columns=-1
        },
        ymin=0,
        ymax=10,
        grid=major,
        axis x line=bottom,
        axis y line=left,
        enlarge x limits=0.03
    ]
    \addplot [mygreen,fill=mygreen!50, bar shift=-6.5pt] table[x=Layer,y=No,col sep=comma] {data/resnet18_4bit_weights.csv};
    \addplot [myred,fill=myred!50, bar shift=-2.2pt] table[x=Layer,y=IO,col sep=comma] {data/resnet18_4bit_weights.csv};
    \addplot [mypurple,fill=mypurple!50, bar shift=+2.2pt] table[x=Layer,y=W,col sep=comma] {data/resnet18_4bit_weights.csv};
    \addplot [myblue,fill=myblue!50, bar shift=+6.5pt] table[x=Layer,y=B,col sep=comma] {data/resnet18_4bit_weights.csv};
    \legend{
        No constraints, 
        Input/output constraints, 
        Weight constraints, 
        Both constraints
    }
    \end{axis}
    \end{tikzpicture}

    \begin{tikzpicture}
    \begin{axis}[
        ybar,
        bar width=4pt,
        width=\textwidth,
        height=3cm,
        ylabel={\shortstack{Activation\\bit width}},
        xlabel={},
        symbolic x coords={Conv1,Conv2,Conv3,Conv4,Conv5,Conv6,Conv7,Conv8,Conv9,Conv10,Conv11,Conv12,Conv13,Conv14,Conv15,Conv16,Conv17,Conv18,Conv19,Conv20,Dense},
        xtick=data,
        x tick label style={rotate=45, anchor=east},
        nodes near coords,
        every node near coord/.append style={font=\small},
        legend style={
            at={(0.5,1.5)},
            anchor=north,
            legend columns=-1
        },
        ymin=0,
        ymax=10,
        grid=major,
        xmajorgrids=false,
        axis x line=bottom,
        axis y line=left,
        enlarge x limits=0.03
    ]
    \addplot [mygreen,fill=mygreen!50, bar shift=-6.5pt] table[x=Layer,y=No,col sep=comma] {data/resnet18_4bit_activations.csv};
    \addplot [myred,fill=myred!50, bar shift=-2.2pt] table[x=Layer,y=IO,col sep=comma] {data/resnet18_4bit_activations.csv};
    \addplot [mypurple,fill=mypurple!50, bar shift=+2.2pt] table[x=Layer,y=W,col sep=comma] {data/resnet18_4bit_activations.csv};
    \addplot [myblue,fill=myblue!50, bar shift=+6.5pt] table[x=Layer,y=B,col sep=comma] {data/resnet18_4bit_activations.csv};
    \end{axis}
    \end{tikzpicture}
    \vspace{-0.7cm}

    \caption{Final activation and weight bit widths of ResNet-18 layers under different constraints for $r_{\mathrm{cell}}=\SI{4}{\bit}$.}
    \label{fig:results_bitwidths}
    \vspace{-0.4cm}
\end{figure*}
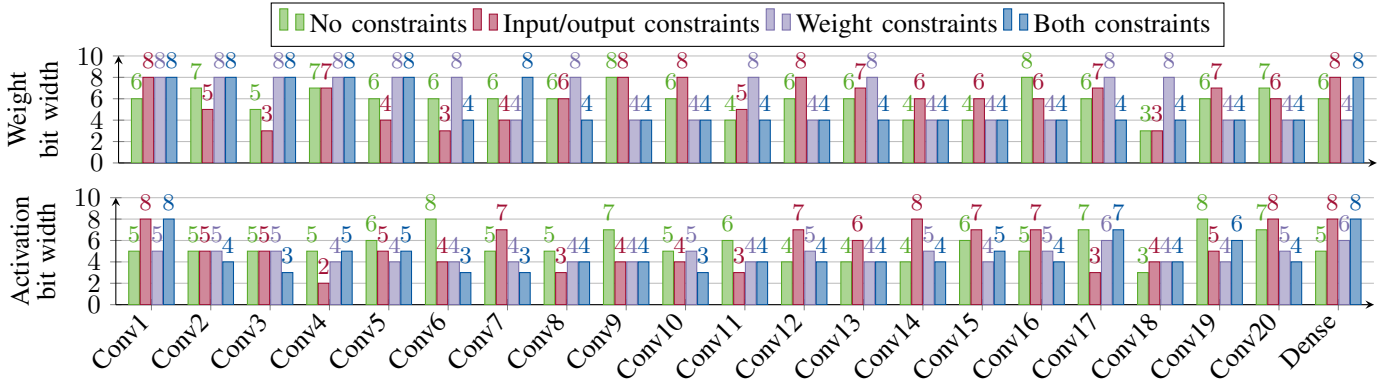

\balance
\subsection{CIM-AQ - Evaluation of Constraints}
Finding optimal \ac{mpq} parameters using \ac{cim-aq} remains time-consuming.
This is why we first explore how different constraint strategies affect the quality of the \ac{mpq} parameters.
By restricting the search space, constraints can help achieve better results.
We evaluate the following constraints:
\begin{itemize}
\item \textbf{Input/output} constraint: The activations and weights in the first and last layer remain at \SI{8}{\bit}, since these layers are typically more sensitive to quantization noise~\cite{lee2022quantune}.
\item \textbf{Weight} constraint: The bit widths of the weights are restricted to multiples of the cell resolution $r_{\mathrm{cell}}$.
\item \textbf{Both} constraints: Both previous constraints.
\end{itemize}

In all subsequent experiments, CIM-AQ is run for 600 episodes,
with 3 QAT epochs per episode (see \Cref{sec:haq_framework}),
on a reduced ImageNet100 dataset with 20,000 training and 10,000 validation images.
The accuracy threshold $acc_\mathrm{loss}$ is set to \SI{5}{\percent} (see \Cref{sec:implementation}).
Once the optimal \ac{mpq} parameters are determined, the \ac{nn} is further trained for 30 epochs on the full ImageNet dataset.
All reported accuracies correspond to the validation accuracy of this final training step.

\Cref{fig:results_bitwidths} shows the final \ac{mpq} bit widths of the \mbox{ResNet-18} layers for weights and activations.
The cell resolution is set to $r_{\mathrm{cell}} = \SI{4}{\bit}$, meaning that weights exceeding \SI{4}{\bit} are distributed across multiple cells.
Although activations are generally considered more sensitive to quantization when using conventional hardware~\cite{zhou2016dorefa},
the results indicate that, for \ac{cim} architectures, activations are in most cases quantized more aggressively than weights.
The reward function (see \Cref{eq:reward_function}) explains this behavior because it explicitly rewards low latency.
The bit width of the activations has a direct impact on latency:
each saved activation bit reduces the \ac{mvm} latency by one cycle.
In contrast, weights affect latency more indirectly:
larger bit widths increase the number of weight bit slices,
which may no longer fit on the crossbar and can lead to additional crossbar writes and compute cycles.

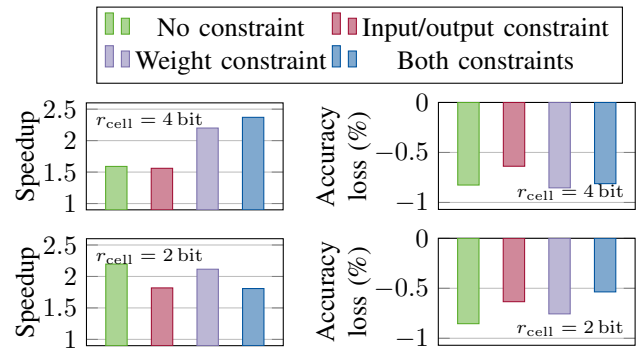
\begin{figure}[!b]
    \vspace{-0.4cm}
    \centering
    \hspace{0.14cm}
    \begin{subfigure}[b]{0.23\textwidth}
        \centering
        \begin{tikzpicture}
        \begin{axis}[
            ybar,
            bar width=8pt,
            width=\textwidth,
            height=3cm,
            symbolic x coords={Speedup},
            ylabel={Speedup},
            ylabel style={yshift=-0.2cm},
            xlabel={},
            xtick=\empty,
            enlarge x limits=0.2,
            grid=major,
            legend style={at={(1.35,1.9)}, legend columns=2, anchor=north, xshift=0.1cm},
            ymin=0.9,
            ymax=2.6
        ]
        \addplot [mygreen,fill=mygreen!50, bar shift=-0.9cm]  table[x=Case,y=No,col sep=comma] {data/resnet18_4bit_speedup.csv};
        \addplot [myred,fill=myred!50, bar shift=-0.3cm]  table[x=Case,y=IO,col sep=comma] {data/resnet18_4bit_speedup.csv};
        \addplot [mypurple,fill=mypurple!50, bar shift=+0.3cm]  table[x=Case,y=W,col sep=comma] {data/resnet18_4bit_speedup.csv};
        \addplot [myblue,fill=myblue!50, bar shift=+0.9cm]  table[x=Case,y=B,col sep=comma] {data/resnet18_4bit_speedup.csv};
        \node[anchor=north west] at (rel axis cs:0,1) {\scriptsize $r_{\mathrm{cell}}=\SI{4}{\bit}$};
        \legend{No constraint, Input/output constraint, Weight constraint, Both constraints}
        \end{axis}
        \end{tikzpicture}
    \end{subfigure}
    \hspace{-0.46cm}
    \begin{subfigure}[b]{0.23\textwidth}
        \centering
        \begin{tikzpicture}
        \begin{axis}[
            ybar,
            bar width=8pt,
            width=\textwidth,
            height=3cm,
            symbolic x coords={Accloss},
            ylabel={\shortstack{Accuracy\\loss (\%)}},
            ylabel style={yshift=-0.2cm},
            xlabel={},
            xtick=\empty,
            enlarge x limits=0.2,
            grid=major,
            ymax=0,
            ymin=-1.07
        ]
        \addplot [mygreen,fill=mygreen!50, bar shift=-0.9cm]  table[x=Case,y=No,col sep=comma] {data/resnet18_4bit_accloss.csv};
        \addplot [myred,fill=myred!50, bar shift=-0.3cm]  table[x=Case,y=IO,col sep=comma] {data/resnet18_4bit_accloss.csv};
        \addplot [mypurple,fill=mypurple!50, bar shift=+0.3cm]  table[x=Case,y=W,col sep=comma] {data/resnet18_4bit_accloss.csv};
        \addplot [myblue,fill=myblue!50, bar shift=+0.9cm]  table[x=Case,y=B,col sep=comma] {data/resnet18_4bit_accloss.csv};
        \node[anchor=south west] at (rel axis cs:0.35,0) {\scriptsize $r_{\mathrm{cell}}=\SI{4}{\bit}$};
        \end{axis}
        \end{tikzpicture}
    \end{subfigure}

    \begin{subfigure}[b]{0.23\textwidth}
        \centering
        \begin{tikzpicture}
        \begin{axis}[
            ybar,
            bar width=8pt,
            width=\textwidth,
            height=3cm,
            symbolic x coords={Speedup},
            ylabel={Speedup},
            ylabel style={yshift=-0.2cm},
            xlabel={},
            xtick=\empty,
            enlarge x limits=0.2,
            grid=major,
            legend style={at={(1.35,1.6)}, legend columns=2, anchor=north},
            ymin=0.9,
            ymax=2.6
        ]
        \addplot [mygreen,fill=mygreen!50, bar shift=-0.9cm]  table[x=Case,y=No,col sep=comma] {data/resnet18_2bit_speedup.csv};
        \addplot [myred,fill=myred!50, bar shift=-0.3cm]  table[x=Case,y=IO,col sep=comma] {data/resnet18_2bit_speedup.csv};
        \addplot [mypurple,fill=mypurple!50, bar shift=+0.3cm]  table[x=Case,y=W,col sep=comma] {data/resnet18_2bit_speedup.csv};
        \addplot [myblue,fill=myblue!50, bar shift=+0.9cm]  table[x=Case,y=B,col sep=comma] {data/resnet18_2bit_speedup.csv};
        \node[anchor=north west] at (rel axis cs:0,1) {\scriptsize $r_{\mathrm{cell}}=\SI{2}{\bit}$};
        \end{axis}
        \end{tikzpicture}
    \end{subfigure}
    \hspace{-0.2cm}
    \begin{subfigure}[b]{0.23\textwidth}
        \centering
        \begin{tikzpicture}
        \begin{axis}[
            ybar,
            bar width=8pt,
            width=\textwidth,
            height=3cm,
            symbolic x coords={Accloss},
            ylabel={\shortstack{Accuracy\\loss (\%)}},
            ylabel style={yshift=-0.2cm},
            xlabel={},
            xtick=\empty,
            enlarge x limits=0.2,
            grid=major,
            legend style={at={(0.5,1.5)}, legend columns=2, anchor=north},
            ymax=0,
            ymin=-1.07
        ]
        \addplot [mygreen,fill=mygreen!50, bar shift=-0.9cm]  table[x=Case,y=No,col sep=comma] {data/resnet18_2bit_accloss.csv};
        \addplot [myred,fill=myred!50, bar shift=-0.3cm]  table[x=Case,y=IO,col sep=comma] {data/resnet18_2bit_accloss.csv};
        \addplot [mypurple,fill=mypurple!50, bar shift=+0.3cm]  table[x=Case,y=W,col sep=comma] {data/resnet18_2bit_accloss.csv};
        \addplot [myblue,fill=myblue!50, bar shift=+0.9cm]  table[x=Case,y=B,col sep=comma] {data/resnet18_2bit_accloss.csv};
        \node[anchor=south west] at (rel axis cs:0.35,0) {\scriptsize $r_{\mathrm{cell}}=\SI{2}{\bit}$};
        \end{axis}
        \end{tikzpicture}
    \end{subfigure}
    \caption{Speedup and top-1 accuracy loss for ResNet-18 in comparison to the \SI{8}{\bit} baseline for different cell resolutions.}
    \label{fig:speedup_drop_resnet}
    \vspace{-0.3cm}
\end{figure}

Another observation is that the activations of the first half of the Conv2D layers are quantized more aggressively than those of the second half.
These layers use fewer weights and thus require fewer crossbar write operations.
At the same time, they perform more \acp{mvm} than later Conv2D layers.
Reducing activation bit width therefore speeds up the front layers more.

To compare the constraint strategies, we analyze their effect on the speedup and accuracy loss trade-off.
Both values are given relative to the \SI{8}{\bit} baseline used in other CIM compilers.
\Cref{fig:speedup_drop_resnet} shows the results for ResNet-18.
The top row uses \SI{4}{\bit} cells, the bottom row \SI{2}{\bit} cells.
Accuracy loss is similar for both resolutions, but speedup is slightly higher for \SI{2}{\bit} cells in all cases except when both constraints are applied.
This is plausible because the latency also depends on the bit width of the weights:
for $r_{\mathrm{cell}}=\SI{4}{\bit}$,
only a weight bit reduction from \SI{8}{\bit} to \SI{4}{\bit} improves the latency,
whereas for \mbox{$r_{\mathrm{cell}}=\SI{2}{\bit}$},
each reduction of \SI{2}{\bit} increases the speedup.

For \SI{4}{\bit} cells, the highest speedups occur with weight constraints or both constraints, reaching over $2.20\times$.
For \SI{2}{\bit} cells, the best speedups are achieved without constraints or with weight constraints.
Accuracy loss, however, is lower with input/output or both constraints, in the best case only \SI{-0.536}{\percent}.
To capture this trade-off, we define the $S/AL$ score as the ratio of speedup to absolute accuracy loss:

\vspace{-0.2cm}
\begin{align}
    S/AL \coloneq \frac{\text{Speedup compared to \SI{8}{\bit} NN}}{|\text{Accuracy loss compared to \SI{8}{\bit} NN}|}
\end{align}
\vspace{0.0cm}

\Cref{tab:resnet_results} lists the $S/AL$ scores for ResNet-18 with \SI{2}{\bit} and \SI{4}{\bit} cells.
Without constraints, the score is lowest.
Applying both constraints yields the highest score.
This confirms that constraints help to reduce the search space and lead to finding better quantization parameters.

\begin{table}[!h]
    \centering
    \footnotesize
    \caption{\label{tab:resnet_results}ResNet18: $S/AL$ score in $\frac{1}{\si{\percent}}$ (higher is better).}
    \begin{tabular}{l|llll}
          & No constraint & Input/Output & Weight & Both \\ \hline
    2 bit & 2.573 (worst) & 2.867 & 2.798 & 3.373 (best) \\
    4 bit & 1.920 (worst) & 2.884 & 2.575 & 2.924 (best)    
    \end{tabular}
    \vspace{-0.1cm}
\end{table}

\begin{figure*}[!tb]
    \centering
    \begin{tikzpicture}
    \begin{axis}[
        ybar,
        bar width=4pt,
        width=\textwidth,
        height=3cm,
        ylabel={Latency},
        xtick=\empty,
        xticklabel style={rotate=90, anchor=east, font=\scriptsize},
        axis line style={-},           
        axis x line=bottom,             
        axis y line=left,               
        tick align=outside,             
        grid=both,
        minor grid style={gray!50, dotted},
        minor y tick num=1,
        xmajorgrids=false,
        enlarge x limits=0.02,
        ymin=0
    ]
    \addplot [myblue,fill=myblue!50] table[x=index,y=value,col sep=comma] {data/vit_latency.csv};
    \end{axis}
    \end{tikzpicture}

    \begin{tikzpicture}
    \begin{axis}[
        ybar,
        bar width=4pt,
        width=\textwidth,
        height=2.75cm,
        ylabel={Speedup},
        xtick=data,
        xticklabel style={rotate=90, anchor=east, font=\scriptsize},
        axis line style={-},           
        axis x line=bottom,             
        axis y line=left,               
        tick align=outside,             
        grid=both,
        minor grid style={gray!50, dotted},
        minor y tick num=1,
        xmajorgrids=false,
        ymin=0,
        enlarge x limits=0.02
    ]
    \addplot [myblue,fill=myblue!50] table[x=index,y=value,col sep=comma] {data/vit_speedup.csv};
    \end{axis}
    \end{tikzpicture}
    \caption{Latency (in s) for \ac{mpq} model and speedup against the \SI{8}{\bit} baseline for each layer of ViT-B/32 with $r_{\mathrm{cell}}=\SI{4}{\bit}$.}
    \label{fig:speedup_drop_vit}
    \vspace{-0.4cm}
\end{figure*}
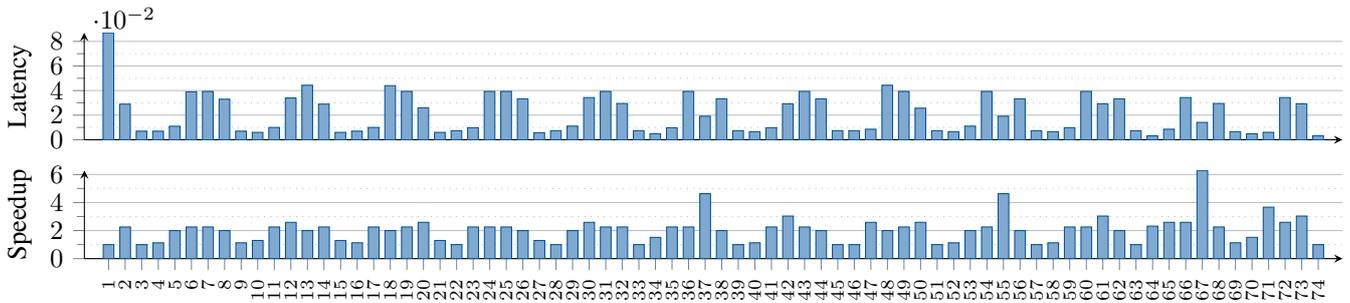

\vspace{-0.1cm}
\subsection{Comparison Against Related Work}
Finally, we evaluate our approach against state-of-the-art \SI{8}{\bit} solutions discussed in \Cref{sec:related_work}.
Since most of the \SI{8}{\bit} compilers operate at a \SI{4}{\bit} cell resolution, our comparison focuses on this configuration.
Moreover, we use both constraints in the following because this strategy delivered the best results in the previous experiments.
Direct, one-to-one comparisons are not possible because many compilers are either closed source or incapable of executing complete \acp{nn}.
To address this, we use \SI{8}{\bit} workloads on our own compiler as a proxy reference while adopting the key compiler optimizations described in~\cite{siemieniuk2021occ}.
Specifically, we assume maximal weight reuse, which is considered the most effective strategy for minimizing costly crossbar rewrites (see \Cref{sec:cim_compiler}).

First, we analyze the results presented in \Cref{fig:speedup_drop_vit},
which illustrate the execution latency and speedup against the \SI{8}{\bit} baseline for each individual layer of the ViT-B/32 model.
The speedup shows that the latency is consistently reduced across most layers.
Due to the applied constraints, the first layer is not quantized beyond \SI{8}{\bit}.
Here, Dense layers have higher latency than MatMul layers,
so their weights are reduced to \SI{4}{\bit}, while most MatMul layers remain at \SI{8}{\bit}.
Dense layers performing dimensionality reduction in the \ac{mlp} block dominate latency due to their large number of weights.
In the second half of the model, their inputs can be reduced to \SI{3}{\bit},
explaining the high speedups in layers $37$, $55$, and $67$.

For all benchmarks, our \ac{mpq} models deliver substantial speedups over the \SI{8}{\bit} baseline.
The results are summarized in \Cref{tab:vit_vgg_resnet_results}.
For VGG-16, our approach yields the highest speedup of $2.48\times$, with an accuracy loss of only \SI{0.086}{\percent}.
Across all \acp{nn}, we consistently achieve speedups of at least $2.20\times$, while the maximum accuracy degradation does not exceed \SI{2.140}{\percent}.

\begin{table}[!h]
    \centering
    \footnotesize
    \caption{\label{tab:vit_vgg_resnet_results}Results: Comparison against \SI{8}{\bit} execution.}
    \vspace{-0.1cm}
    \begin{tabular}{c|S[table-format=1.3]S[table-format=-1.3]S[table-format=2.3]}
          & {Speedup} & {Accuracy change (\unit{\percent})} & {$S/AL$ score (\unit{$\frac{1}{\percent}$})} \\ \hline
    ResNet-18 & 2.37 & -0.812 &  2.924 \\
    VGG-16    & 2.48 & -0.086 & 28.862 \\
    ViT-B/32  & 2.20 & -2.140 &  1.028 \\
    \end{tabular}
    \vspace{-0.3cm}
\end{table}

\section{Conclusion}
In this work, we introduced a \ac{mpq} training and compilation framework for \ac{rram}-based \ac{cim} accelerators.
We first extended and improved the \ac{haq} framework to create CIM-AQ,
a reinforcement learning-based optimizer that explores the large \ac{mpq} search space for CIM architectures.
CIM-AQ balances latency and accuracy.
We also introduced constraint strategies to further improve search efficiency.
Our experiments with ResNet-18, VGG-16, and ViT-B/32 on ImageNet demonstrate that, in the best case,
the proposed approach achieves up to a $2.48\times$ speedup over existing \SI{8}{\bit} compilers,
with an accuracy loss of only \SI{0.086}{\percent}.
These results confirm that our \ac{mpq} framework can significantly reduce inference latency on \ac{cim} while preserving high accuracy.

At this stage, the observed accuracy losses stem solely from quantization effects.
Future work will extend the framework to also include crossbar non-idealities in the training process.
\vspace{-0.345cm}

\section*{Acknowledgments}
{\small This work was funded by the Federal Ministry of Research, Technology and Space
in the projects NeuroSys II (03ZU2106CA), NEUROTEC-II (16ME0399), and EXIST (03EFWNW338).}

\bibliographystyle{IEEEtran}
\bibliography{bibtexentry}

\end{document}